\documentclass{article}

\usepackage{arxiv}

\usepackage[utf8]{inputenc} % allow utf-8 input
\usepackage[T1]{fontenc}    % use 8-bit T1 fonts
\usepackage{hyperref}       % hyperlinks
\usepackage{url}            % simple URL typesetting
\usepackage{booktabs}       % professional-quality tables
\usepackage{amsfonts}       % blackboard math symbols
\usepackage{nicefrac}       % compact symbols for 1/2, etc.
\usepackage{microtype}      % microtypography
\usepackage{listings}
\usepackage{color}
\usepackage{subcaption}

\usepackage{times}
\usepackage{epsfig}
\usepackage{graphicx}
\usepackage{amsmath}
\usepackage{amssymb}
\usepackage{listings}
\usepackage{url}
\usepackage{footmisc}

\title{Kannada-MNIST: A new handwritten digits dataset for the Kannada language}

\author{
  Vinay Uday Prabhu \\
  \\  \texttt{ dig.mnist@gmail.com}
}

\begin{document}
\maketitle

\begin{abstract}
 In this paper, we disseminate a new handwritten digits-dataset, termed \textit{Kannada-MNIST}, for the Kannada script, that can potentially serve as a direct drop-in replacement for the original MNIST dataset\cite{mnist}. In addition to this dataset, we  
disseminate an additional real world handwritten dataset (with $10k$ images),  which we term as the \textit{Dig-MNIST}\footnote{The word \textit{dig} is an ode to the diminutive used to denote Kannada speakers in the main author's alma mater \cite{richter2006student}} dataset that can serve as an out-of-domain test dataset. We also duly open source all the code as well as the raw scanned images along with the scanner settings so that researchers who want to try out different signal processing pipelines can perform end-to-end comparisons. We provide high level morphological comparisons with the MNIST dataset and provide baselines accuracies for the dataset disseminated. The initial baselines\footnote{The companion github repository for this paper is : \url{https://github.com/vinayprabhu/Kannada_MNIST}} obtained using an oft-used CNN architecture ($96.8\%$ for the \textit{main test-set} and $76.1\%$ for the \textit{Dig-MNIST} test-set) indicate that these datasets do provide a sterner challenge with regards to generalizability than MNIST or the KMNIST datasets. We also hope this dissemination will spur the creation of similar datasets for all the languages that use different symbols for the numeral digits.
\end{abstract}

% keywords can be removed

\section{Introduction}
\begin{figure}[ht]
\begin{center}
\includegraphics[width=0.8\linewidth]{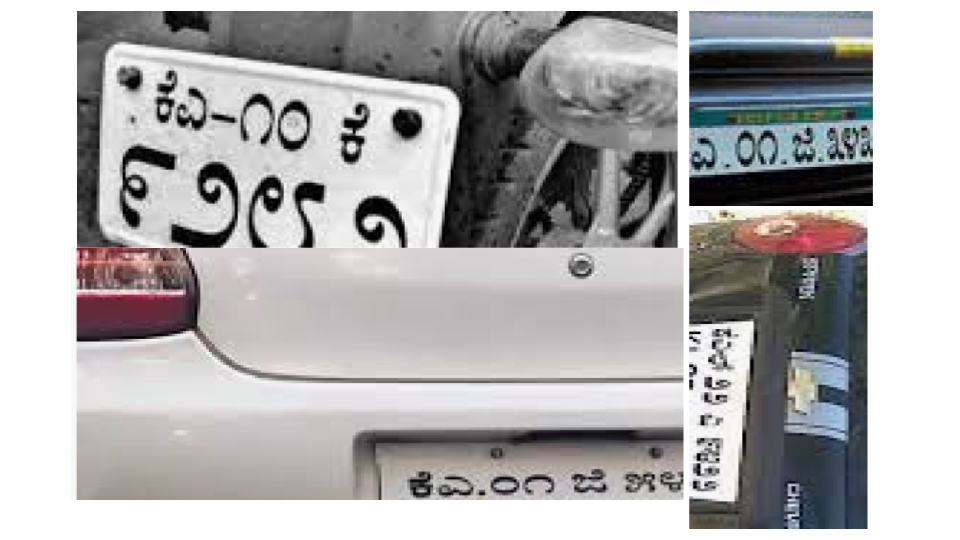}
\end{center}
   \caption{Usage of the Kannada numerals on vehicular license plates}
\label{fig:num_plates}
\end{figure}

Kannada is the official and administrative language of the state of Karnataka in India with nearly 60 million speakers worldwide \cite{cite_1}. Also, as per articles 344(1) and 351 of the Indian Constitution, Kannada holds the status of being one of the 22 scheduled languages of India \cite{cite_2}. The language is written using the official Kannada script, which is an \textit{abugida} of the Brahmic family and traces its origins to the \textit{Kadamba script} (325-550 AD). 
\\ Distinct glyphs are used to represent the numerals 0-9 in the language that appear distinct from the modern Hindu-Arabic numerals in vogue in much of the world today. Unlike some of the other archaic numeral-systems, these numerals are very much used in day-to-day affairs in Karnataka, as in evinced by the prevalence of these glyphs on license-plates of vehicles captured in fig \ref{fig:num_plates}.

\begin{figure}[ht]
\begin{center}
\includegraphics[width=0.8\linewidth]{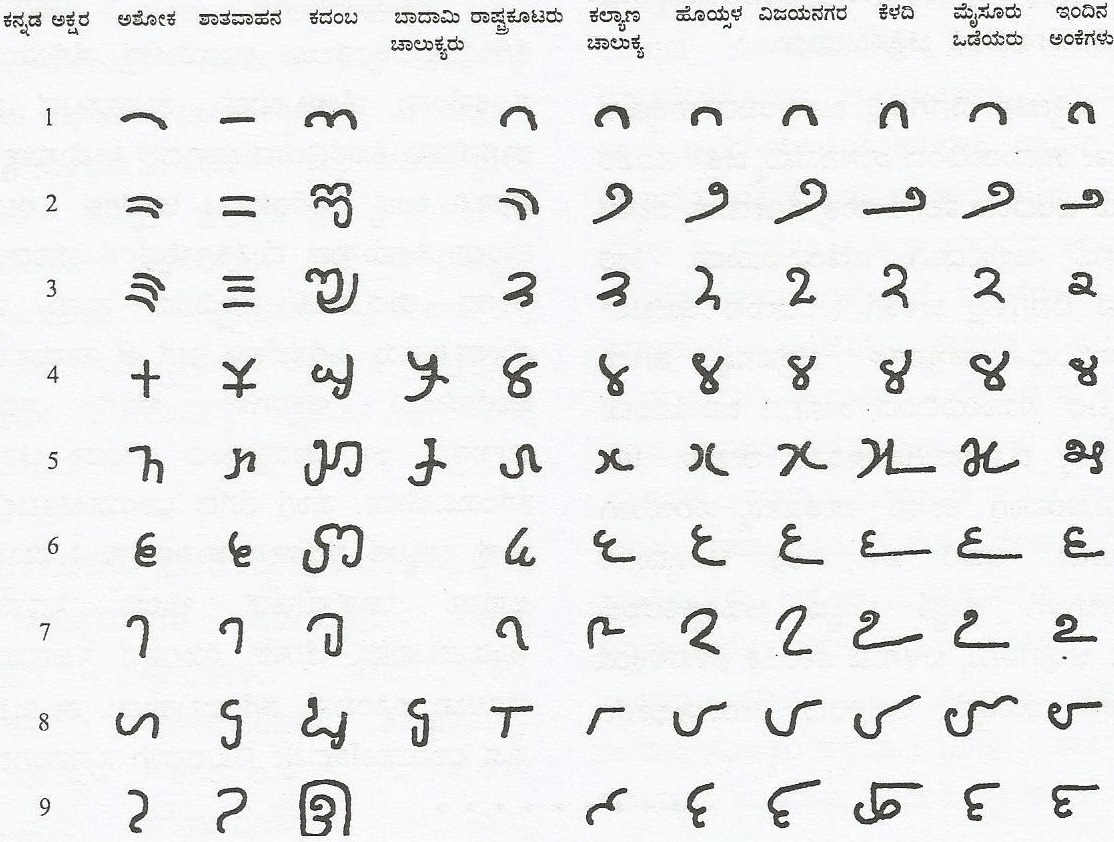}
\end{center}
   \caption{Evolution of the Kannada numerals through the ages (\cite{kannada_lipi_vikasa})}
\label{fig:evolution}
\end{figure}

Fig \ref{fig:evolution} captures the evolution of the numerals through the ages. Modern Kannada scholars \cite{Gudnapur_hindu} posit that the emergence of these numeral-glyphs can be traced to the \textit{Gudnapur inscriptions}  \cite{Gudnapur_inscr}, dating back to the $6^{th}$ century AD when the Kadamba rulers held sway over the region\cite{Gudnapur_paper}. 
The Kannada digits for 0-9 are shown in Fig 1 (Unicode: 0CE6 through to 0CEF) \cite{cite_3} .
\begin{figure}[ht]
\begin{center}
\includegraphics[width=0.8\linewidth]{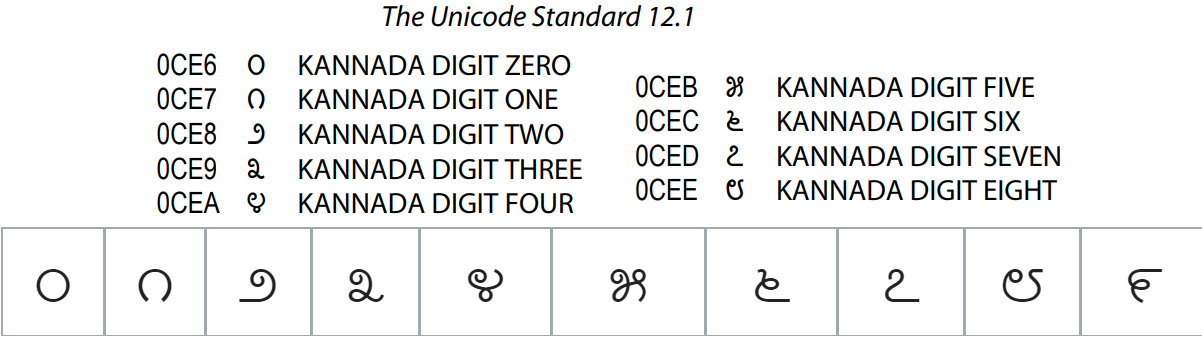}
\end{center}
\caption{The character code tables for Kannada-MNIST from the Unicode Standard, Version 12.1}
\label{fig:unicode}
\end{figure}

\begin{figure}[ht]
\begin{center}
\includegraphics[width=0.8\linewidth]{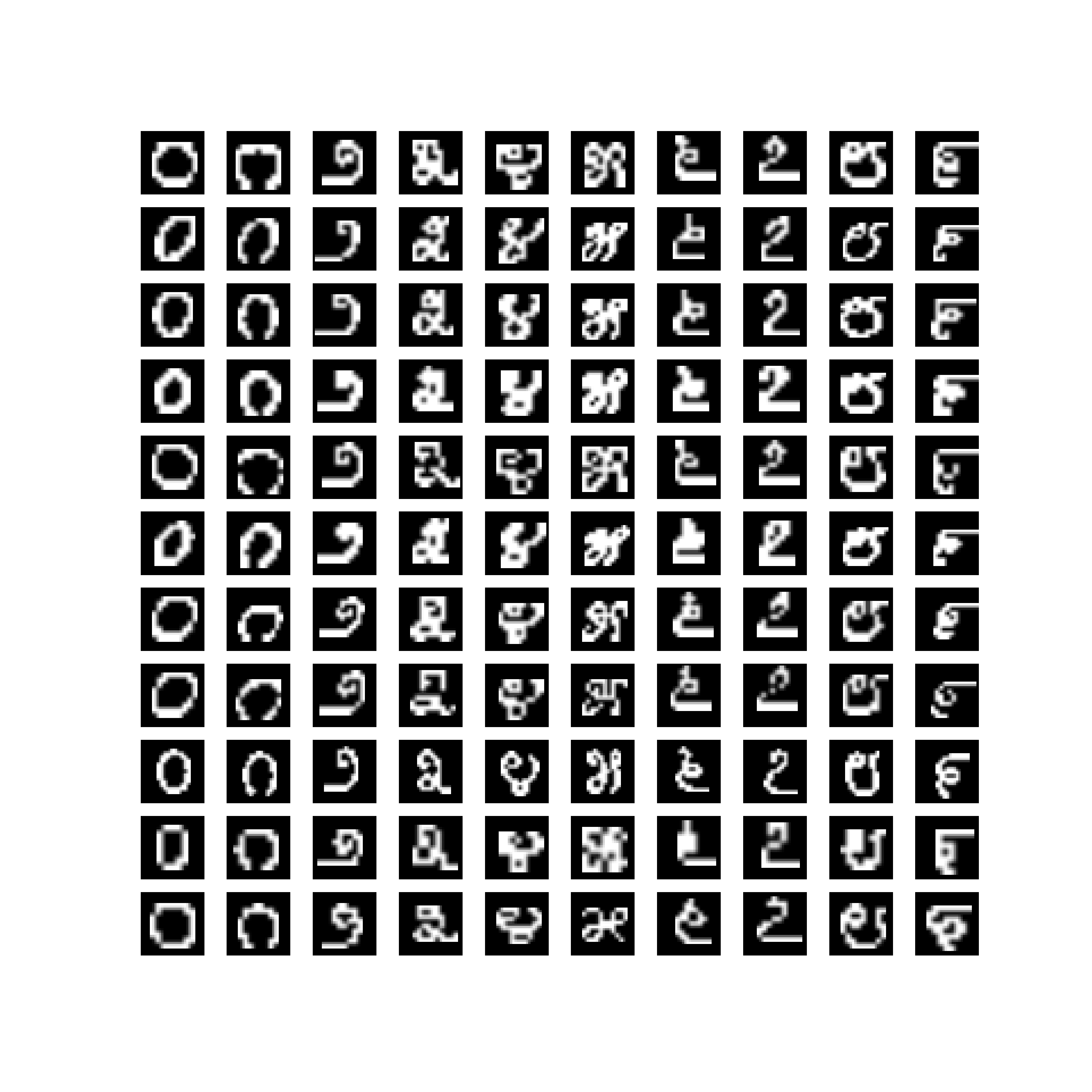}
\end{center}
   \caption{MNIST-ized renderings of the 0-9 Kannada numerals in 11 modern fonts}
\label{fig:all_fonts}
\end{figure}

Fig \ref{fig:all_fonts} captures the MNIST-ized renderings of the variations of the glyphs across the following modern fonts: \textit{ Kedage, 
 Malige-i,
 Malige-n,
 Malige-b,
 Kedage-n,
 Malige-t,
 Kedage-t,
 Kedage-i,
 Lohit-Kannada,
 Sampige and
 Hubballi-Regular}.
 
\subsection{The curious case of glyphs for 3,7 and 6}
\label{subsec:2_6}
\begin{figure}[ht]
\begin{center}
\includegraphics[width=0.8\linewidth]{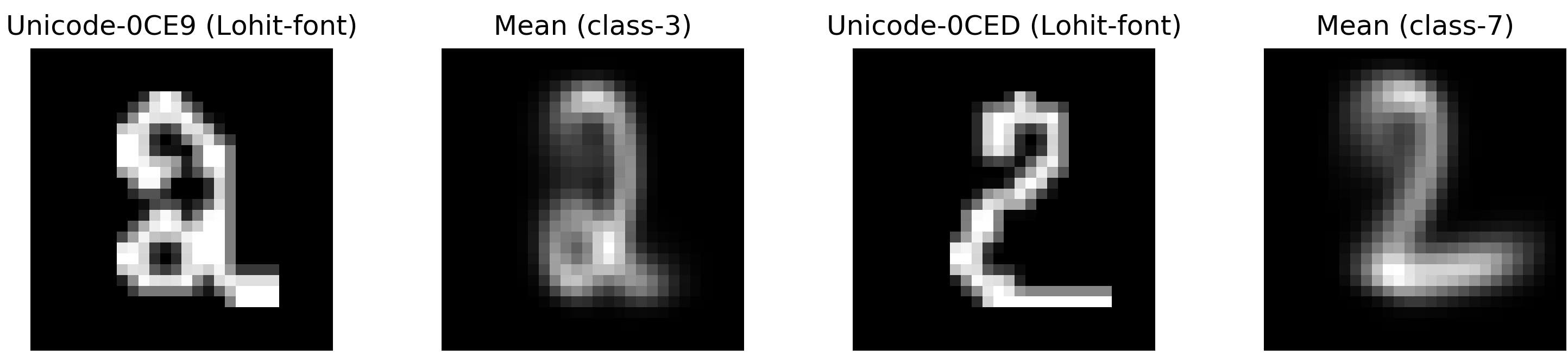}
\end{center}
\caption{The similarity between the glyphs for 3 and 7 in Kannada}
\label{fig:3_7}
\end{figure}
In this section, we focus on some idiosyncrasies with regards to the shapes of the glyphs used to represent numerals in Kannada. Three interesting observations emerge from Fig\ref{fig:unicode} and Fig\ref{fig:all_fonts}. 
\begin{figure}[ht]
\begin{center}
\includegraphics[width=0.8\linewidth]{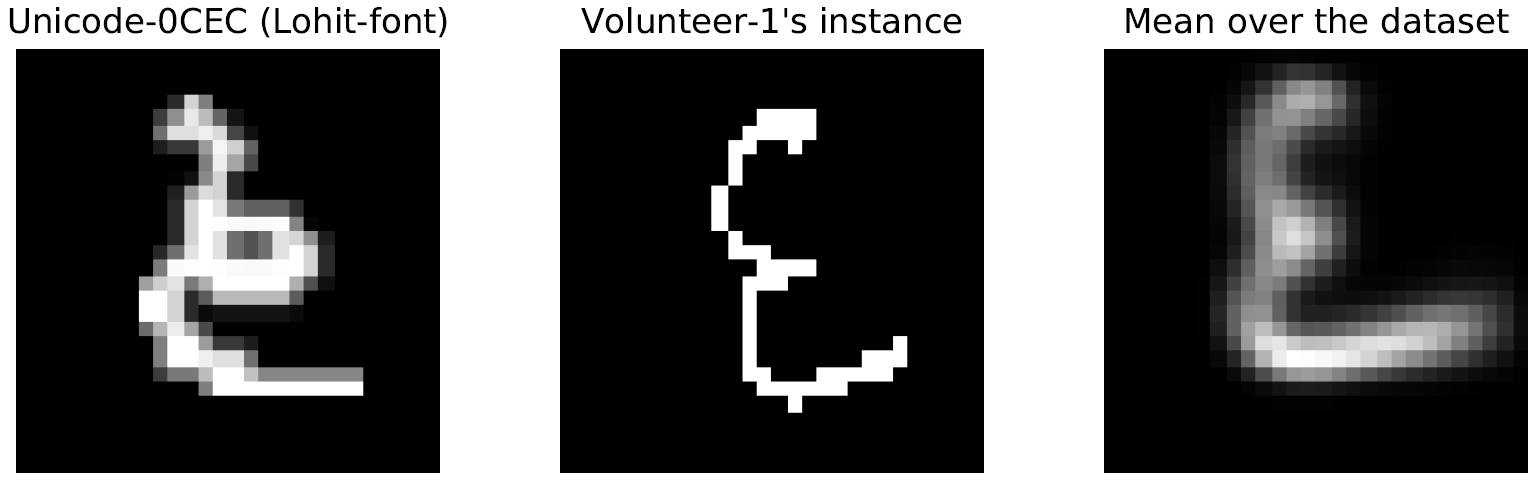}
\end{center}
\caption{Deformations in the numeral glyphs for 6}
\label{fig:six}
\end{figure}

The first observation is that the glyph for 0 is the same as in the Hindu-Arabic system. Secondly, the shapes of the digits for 3 and 7 in Kannada look rather similar to the glyph for 2 in the modern Hindu-Arabic numeral system (Fig \ref{fig:3_7}).

These will be leveraged during our dataset curation procedure as a sanity check for scanned and segmented digits using a pre-trained MNIST-digits classifier.
%These two observations will be used to perform sanity-checks while parsing the digits from the handwritten scans. 
\\ The third observation is related to the peculiar intra-class variation of the representation (refer to fig \ref{fig:all_fonts} and fig \ref{fig:evolution}) for 6 across different fonts and different eras. The modern day deformations are represented in Fig \ref{fig:six}, where we observe the deviation from the puritanical \textit{textbook} representation of the symbol (as seen in the unicode-derived image on the extreme left) and the more colloquial usage which looks like a mirror image of 3 in the Hindu-Arabic system. As will be seen in the upcoming section, many of the volunteers who helped curate the dataset used one or both of these glyphs, resulting in high intra-class variation.
\subsection{Related work}
There have been some nascent attempts made towards Kannada handwritten digit classification, albeit at a smaller scale. In \cite{2006recognition}, the authors used the chain code histogram idea to achieve 98\% accuracy on a dataset of 2300 digit-images. In \cite{2007recognition}, the authors used a nearest neighbor classifier to achieve 91\% accuracy of 250 test numerals. Support Vector Machines (SVMs) were used to achieve 98\% accuracy on a small dataset of 5000 $40 \times 40$ numeral-images in \cite{2010printedrajput}. The largest dataset currently used in academic literature that contains Kannada characters is the \texttt{Chars74k dataset} \cite{chars74k} that contains 657 characters of the
Kannada script collected using a tablet PC, albeit  with a mere 25 samples per-number. In \cite{ganesh2016deep}, the authors harnessed standard augmentation techniques to create an augmented dataset of 18000 digit images harnessing the \texttt{Chars74k dataset} and trained Convolutional Neural Networks (CNNs) and Deep Belief Networks (DBNs) to obtain $98\%$ test accuracy. Earlier this year, we proposed a Seed-Augment-Train/Transfer (SAT) framework that contains a synthetic seed image dataset generation procedure for languages with different numeral systems using freely available open font file datasets (Lohit to be more specific). This seed dataset of images was then augmented to create a purely synthetic training dataset, using which we trained a deep neural network and tested on held-out real world small-sized handwritten digits dataset spanning five Indic scripts, Kannada, Tamil, Gujarati, Malayalam, and Devanagari, containing 1280 digits each.
\\Through this paper, we hope to address this paucity of an MNIST-sized dataset for the Kannada language.
\subsection{Main contributions of the paper}
The main contributions are:
\begin{enumerate}
\item Contributing a real world handwritten \textit{Kannada-MNIST} dataset that was collected in Bangalore, India, that can potentially serve as a direct drop-in replacement for the original MNIST dataset\cite{mnist}. 
\item Contributing an additional 10k real world handwritten \textit{Dig-MNIST} dataset that was collected in Redwood City, CA, that can serve as an out-of-domain test dataset.
\item Open sourcing all the code required to generate such datasets for other languages. 
\item Open-sourcing the raw scanned images along with the scanner settings so that researchers who want to try out different signal processing pipelines can perform end-to-end comparisons.
\item Performing high level morphological comparisons with the MNIST dataset and providing baselines accuracies for the dataset disseminated.
\item Open sourcing the code-templates for generating synthetic seed images for various modern Kannada fonts.
\end{enumerate}

The rest of the paper is organized as follows: Section-2 covers the dataset preparation process, Section-3 details the comparisons vis-a-vis the standard MNIST dataset. In Section-4, we present the classification baseline results obtained using an off-the-shelf CNN,  and Section-5 concludes the paper.

\section{Dataset Creation}
In order to avoid the kind of uncertainties, folklore and trivia surrounding MNIST (as evinced  in \cite{qmnistpaper}), we have decided to detail and open source all aspects of the data collection process. Further, we have also decided to open-source the raw scan images to facilitate end-to-end experimentation with disparate signal processing pipelines.
In this section, we will cover the details of creating the following two datasets:
\begin{enumerate}
\item The \textit{main} Kannada-MNIST dataset that consists of a training set of  60000 $28 \times 28$ gray-scale sample images and a test set of 10000 sample images uniformly distributed across the 10 classes. This dataset is based off of the efforts of 65 volunteers from Bangalore, India, who are native speakers and users of the Kannada language and the script. This was curated to serve as a direct one-to-one drop-in replacement for the original MNIST dataset (akin to Fashion-MNIST \cite{fashion_mnist} and K-MNIST \cite{k_mnist} datasets).
\item The \texttt{Dig-MNIST} dataset that consists of 10240 $28 \times 28$ gray-scale images that was curated with the purpose of providing a more challenging test dataset that was curated in Redwood City, CA, with the help of volunteers many of whom were encountering the Kannada script for the first time and had fair difficulty in replicating the shape of the glyphs. This test dataset, we hope will facilitate domain adaptation experiments.
\end{enumerate}

\begin{figure}[ht]
\begin{center}
\includegraphics[width=\textwidth,height=\textheight,keepaspectratio]{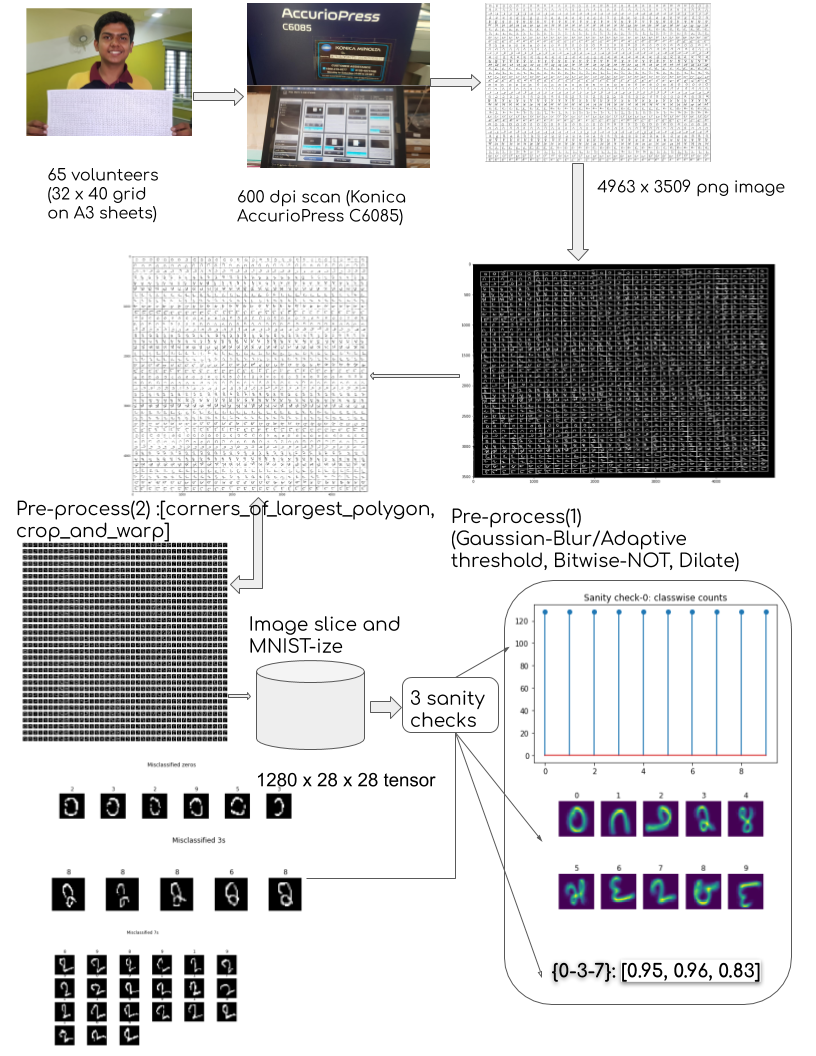}
\end{center}
\caption{The main dataset creation workflow}
\label{fig:dataset_main}
\end{figure}
\subsection{Main dataset}
\label{sec:main_dataset}
Fig \ref{fig:dataset_main} presents the work-flow followed to curate the \textit{main} Kannada-MNIST dataset. The whole process was split into four phases: Data-gathering, pre-processing and slicing, Sanity-check, Train-test split. In the following subsections, we will cover each of these in detail.
\subsubsection{Data-gathering}
65 volunteers were recruited in Bangalore, India, who were native speakers of the language as well as day-to-day users of the numeral script. Each volunteer filled out an A3 sheet containing a  $32 \times 40$  grid. This yielded filled-out A3 sheets containing 128 instances of each number which we posit is large enough to capture most of the natural intra-volunteer variations of the glyph shapes. All of the sheets thus collected were scanned at 600 dots-per-inch resolution using the \texttt{Konica Accurio-Press-C6085} scanner that yielded 65 $4963 \times 3509$ png images.
\subsubsection{Pre-processing and slicing}
In this sub-phase, each of the $4963 \times 3509$ sized scanned $32 \times 40$ grid png images were passed through two pre-processing stages\footnote{\url{https://bit.ly/32WTxeT}} as detailed in \cite{sudoku_extractor}.  This approach was originally used as an extraction framework to eke out the digits for a \textit{Sudoku}-solver. The first pre-processing stage entails:
\begin{enumerate}
\item Applying a Gaussian-blur filter of kernel-size $9 \times 9$
\item Performing adaptive thresholding using 11 nearest neighbour pixels 
\item Applying a \texttt{bitwise-NOT} operator to perform colour inversion to ensure that the target gridlines have non-zero pixel values
\end{enumerate}
The second pre-processing phase entailed two operations. The first was to estimate the corners of the largest polygon that was then harnessed to crop and warp the $32 \times 40$  grid-image. The intermediate images after these phases are as shown in Fig \ref{fig:dataset_main}.
\\The cropped-and-warped image thus obtained was then segmented into 1280 slices to yield the constituent individual digit images which were then \textit{MNIST-ized}\footnote{\url{https://medium.com/@o.kroeger/tensorflow-mnist-and-your-own-handwritten-digits-4d1cd32bbab4}}. For this, we followed the procedure in \cite{mnistize} that entails pixel-thresholding, row-column padding and finally inflicting a \texttt{Best-shift} transformation to drag the current center-of-mass of the digit image to the center of the target \textit{MNIST-ized} $28 \times 28$ image.
At the end of this phase, we had a $128 \times 10 \times 28 \times 28$ image tensor per scanned image. The class-label associated with each image was obtained using the row index of the image in the $32 \times 40$ grid.
\subsubsection{Sanity-check}
\label{sec:sanity_check}
One natural question that emerged during our new dataset curation was how to ensure the MNIST-compatibility of the same? In order to assuage these concerns, we decided to literally use a CNN pre-trained on the original MNIST digits and perform inference on the newly created digit images targeting classes in Kannada that looked \textit{similar} to the MNIST digits.This formed an integral part of a series of \textit{sanity checks} we performed that are as shown in Fig \ref{fig:dataset_main} and listed below:
\begin{enumerate}
\item Firstly, we perform class-wise checks by looking at the histogram of counts of the labels as well as eye-balling out the class-wise mean images of the $128 \times 10$ digits array and visually verifying that they indeed \textit{look} like their archetypal glyphs.
\item Secondly, using the observations made in Section \ref{subsec:2_6}, we perform 3 sets of classifications on the \textit{MNIST-ized} digit images. We use a  high accuracy ($99.4\% $ test-set accuracy) CNN\footnote{\url{https://github.com/keras-team/keras/blob/master/examples/mnist_cnn.py}} pre-trained on MNIST digits to classify the images belonging to class \texttt{zero}(as the glyph for zero is the same), \texttt{three} and \texttt{seven} (as the glyphs look very similar to 2 in MNIST). This produces a triple of accuracies which are used to ascertain the \textit{quality} and \textit{MNIST-likeness} of the images produced by the parsing procedure we've deployed. With regard to Fig \ref{fig:dataset_main}, $95\%$ of the 128 zero-images were classified by the MNIST-CNN as zero, $96\%$ of the 128 \texttt{three}-class images were classified by the MNIST-CNN as class-2 and $83\%$ of the 128 \texttt{seven}-class images were classified by the MNIST-CNN as class-2.
\end{enumerate}
We have duly shared the implementation of the step-wise procedure described above as a \texttt{colab} notebook accessed here: 
\url{https://github.com/vinayprabhu/Kannada_MNIST/blob/master/colab_notebooks/0)Scan_parse_example_main.ipynb}
\subsubsection{Train-test split: Worst cohort selection}
\begin{figure}[t]
\begin{center}
\includegraphics[width=0.8\linewidth]{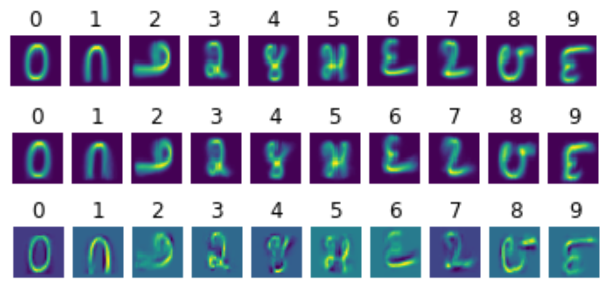}
\end{center}
   \caption{Class-wise mean images of the train set, the test set and the difference between the means of the train and test sets}
\label{fig:train_test_means}
\end{figure}

Using the curation procedure described above, we were able to collect $65\times1280=83200$ images. In order to ensure that our dataset would serve as drop-in replacement to the MNIST dataset, we had to select $60000$ images for the final training set and $10000$ images for the final test dataset. Upon random selection, we were able to hit similar levels of accuracy (>$99\%$) that is achieved for the MNIST dataset. This could very well be attributed to the fact that random sampling based train-test data segmentation essentially allows the CNN to cover the span of the possible \textit{glyph-modes} used by a user to represent a numeral whereas the \textit{real} generalization challenge lies in being able to model the possible deviations that the glyphs shape might take across \textit{unseen} users. Hence, we first sorted the users according to their difficulty scores. The difficulty score for a user was computed by taking the mean of the elements of the proxy-score-vector, which represents the probabilities that the user's 0, 3 and 7 representations in Kannada would be classified as 0, 2 and 2 by the MNIST-CNN classifier as explained in Section \ref{sec:sanity_check}. We then picked the \texttt{top/worst} 8 users into a test cohort and sampled $10000$ digits to form the final test dataset. We then picked the next 47 users and sampled $60000$ digits to form the final train dataset. This implies that any test-accuracy achieved by a machine learning classifier model will actually map to the ability of the classifier to predict the digit-classes from images emanating from users \textit{hitherto unseen} during the training phase. The triple of 0-3-7 class accuracy-vectors of the train and test datasets thus formed were $[0.943, 0.962, 0.9575]$ and $[0.825, 0.87, 0.743]$ respectively.\\We have shared the implementation of this procedure through a \texttt{colab} notebook shared here: \url{https://github.com/vinayprabhu/Kannada_MNIST/blob/master/colab_notebooks/1b)_Main_dataset_tensor_generation_worst_cohort.ipynb}

% \url{https://github.com/vinayprabhu/Kannada_MNIST}
The class-wise mean images of the train set (top-row) , the test set (second-row) and the difference between the train and test classwise-means is shown in Fig \ref{fig:train_test_means}

\subsection{ 10k Dig-MNIST dataset}
As stated above, we also disseminate an additional more challenging 10k \texttt{Dig-MNIST} dataset in this paper that was collected using volunteers in Redwood City, many of whom were, in fact, seeing the numeral glyphs for the first time and trying their best to reproduce the shapes. This sampling-bias, combined with the fact that we used a completely different writing sheet dimension and scanner settings, resulted in a dataset that would turn out to be far more challenging than the \textit{easy} test dataset curated in the above sub-section. The rest of this sub-section details the specifics of the procedure and the companion colab notebook can be obtained at:
\url{https://github.com/vinayprabhu/Kannada_MNIST/blob/master/colab_notebooks/2)_Kannada_MNIST_10k_RWC.ipynb}.
\subsection{Dataset curation}
\begin{figure}[t]
\begin{center}
\includegraphics[width=0.8\linewidth]{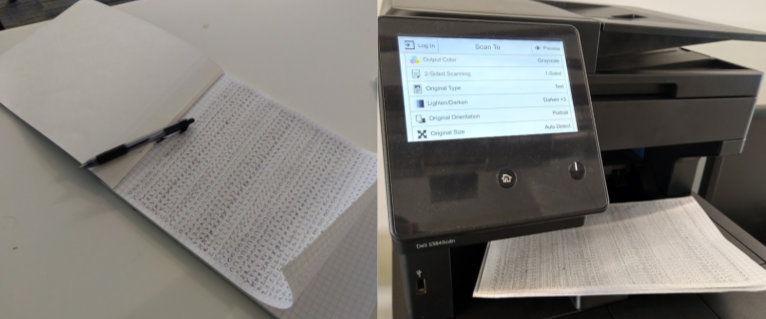}
\end{center}
   \caption{Preparing the \textit{dig}-dataset in Redwood City}
\label{fig:dig_dataset}
\end{figure}

8 volunteers aged 20 to 40 were recruited to generate a $32 \times 40$ grid of Kannada numerals (akin to \ref{sec:main_dataset}), all written with a black ink Z-Grip Series | Zebra Pen on a commercial Mead Cambridge Quad Writing Pad, 8-1/2" x 11", Quad Ruled, White, 80 Sheets/Pad book. We then scan the sheet(s) using a Dell - S3845cdn scanner (See Fig \ref{fig:dig_dataset})with the following settings:
\begin{itemize}
\item Output color: Grayscale
\item Original type: Text
\item Lighten/Darken: Darken+3
\item Size: Auto-detect
\end{itemize}

\begin{figure}[t]
\begin{center}
\includegraphics[width=0.8\linewidth]{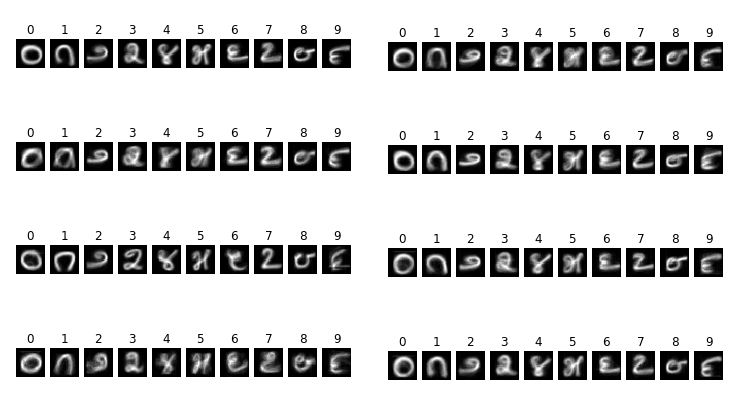}
\end{center}
   \caption{User-wise class-wise mean images for the Dig-10k dataset}
\label{fig:dig-10k-means}
\end{figure}

The reduced size of the sheets used for writing the digits (US-letter vis-a-vis A3) resulted in smaller scan (.tif) images that were all approximately $1600 \times 2000$. The \texttt{Darken+3} scanner option resulted in the grid lines being visible enough that it allowed us to use the same signal processing pipeline detailed in Section \ref{sec:main_dataset} built on the sudoku-digit extraction idea. Fig
\ref{fig:dig-10k-means} captures the user-wise class-wise mean-images for the dataset thus curated.
%%%%%%%%%%%%%%%%%%%%%%%%%%%%%%%%%%%%%%%%%%%%%%%%%%%%%%%%%%%%%%%%%%%%%%%%%

\section{Comparisons with the MNIST dataset}
In the section, we provide some qualitative and quantitative comparisons between the MNIST and the Kannada-MNIST datasets\footnote{In all the results shared in the section, we used the MNIST-10k-Test dataset and the Kannada-MNIST-Test dataset for the experiments.}
\subsection{Morphological comparisons}
\begin{figure}[ht]
\begin{center}
\includegraphics[width=0.8\linewidth]{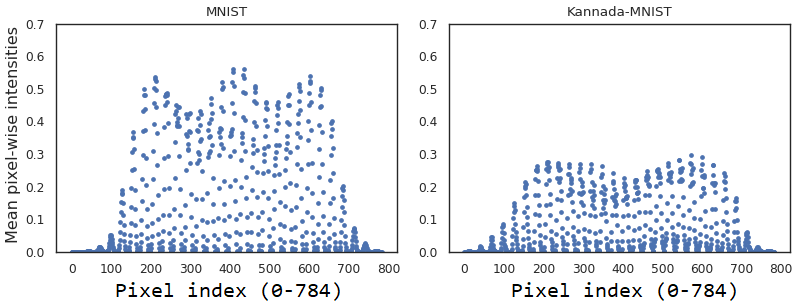}
\end{center}
   \caption{Mean pixel-wise intensities comparisons between MNIST and the Kannada-MNIST datasets}
\label{fig:pixel_int}
\end{figure}

\begin{figure}[ht]
\begin{center}
\includegraphics[width=\textwidth,height=\textheight,keepaspectratio]{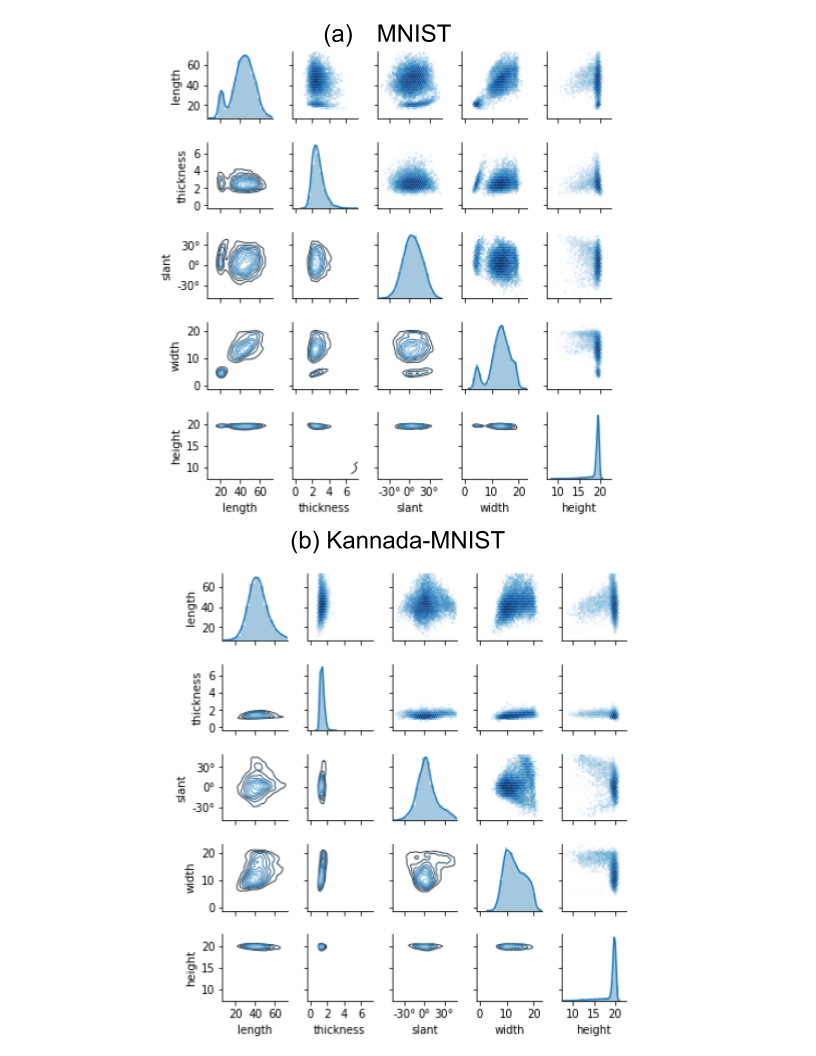}
\end{center}
   \caption{Morphological comparisons between MNIST and the Fashion-MNIST}
\label{fig:morph_both}
\end{figure}

In Fig \ref{fig:pixel_int}, we provide a comparison of the mean pixel-wise intensities between the MNIST and the Kannada-MNIST datasets. As seen, the Kannada-MNIST dataset is much less \textit{peaky} with a maximal mean pixel-intensity of $\sim0.3$ as compared to the MNIST dataset, that has a few pixel indices with mean pixel-intensities of $\sim0.6$. In Fig \ref{fig:morph_both}, we used the \textit{Morpho-MNIST} framework \cite{castro2018morphomnist} to generate the statistics of morphological traits such as length, thickness, slant, width and height of the handwritten digits for the two datasets. As seen, the bi-modality of length as well as width is less pronounced for the Kannada digits. The slant-to-width joint-scatter-plots were visibly different between the two datasets as well.
\subsection{Dimensionality reduction comparisons}
\begin{figure}[ht]
\begin{center}
\includegraphics[width=0.8\linewidth]{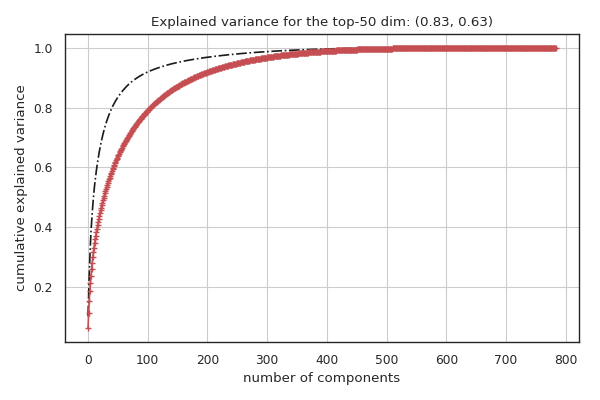}
\end{center}
   \caption{PCA analysis for the two datasets}
\label{fig:pca_expl}
\end{figure}

\begin{figure}[ht]
\begin{center}
\includegraphics[width=1.1\linewidth]{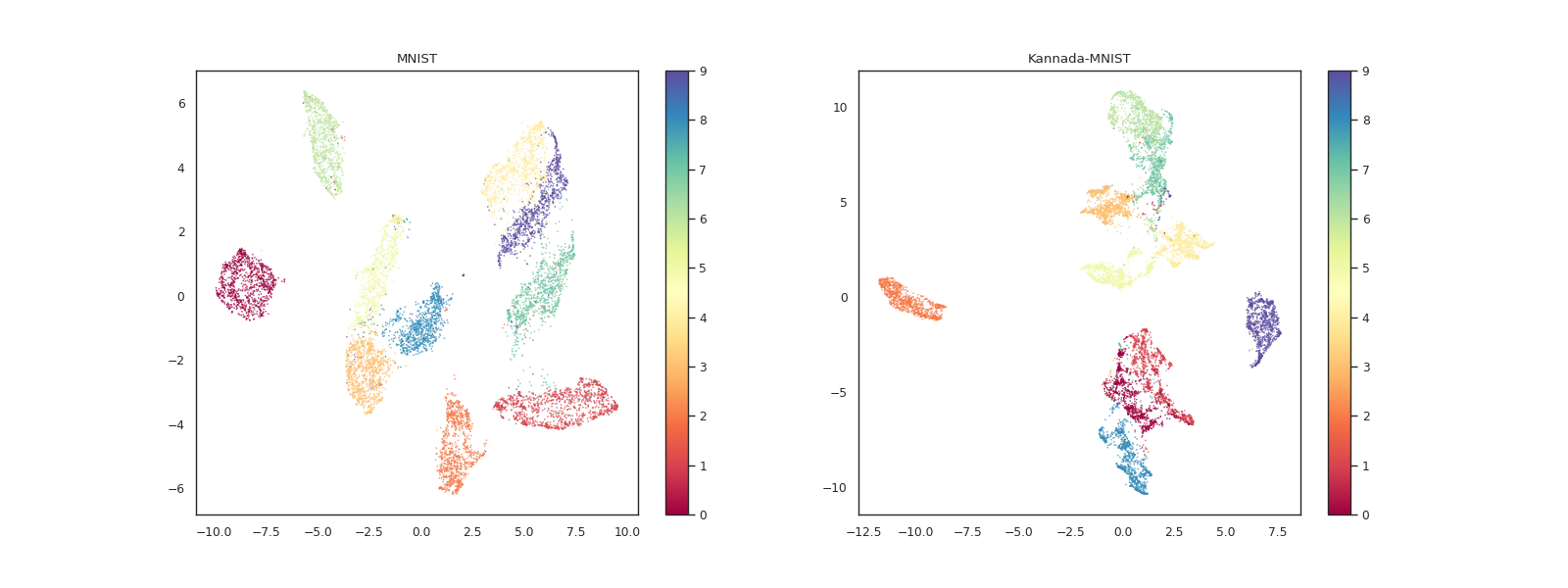}
\end{center}
   \caption{Two-dimensional Uniform Manifold Approximation and Projection (UMAP) plots for the two datasets}
\label{fig:umap_both}
\end{figure}

To begin with, we used the Uniform Manifold Approximation and Projection (UMAP) \cite{mcinnes2018umap} technique to visualize a two-dimensional projection of the two datasets. As seen in Fig \ref{fig:umap_both}, the two sub-plots paint a very different picture of the lower dimensional representations for the two datasets. We also performed dimensionality reduction analysis using PCA to understand the variation of explained variance across the PCA components. As seen in Fig \ref{fig:pca_expl},the top-50 PCA components \textit{explain} $83\%$ of the total variance for the MNIST dataset and only $63\%$ for Kannada-MNIST.


%%%%%%%%%%%%%%%%%%%%%%%%%%%%%%%%%%%%%%%%%%%%%%%%%%%%%%%%%%%%%%%%%%%%%%%%%%%%%%%%
\section{Classification results}
\begin{figure}[t]
\begin{center}
\includegraphics[width=0.8\linewidth]{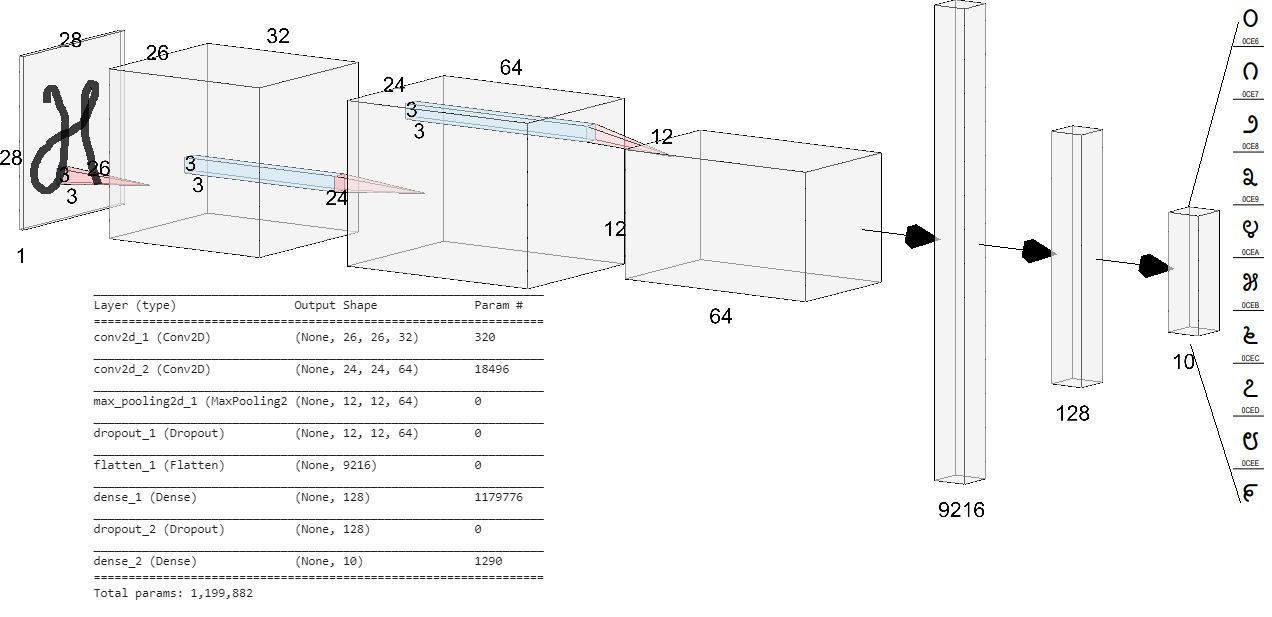}
\end{center}
   \caption{The CNN architecture used in the paper}
\label{fig:cnn}
\end{figure}
\begin{figure}[t]
\begin{center}
\includegraphics[width=0.8\linewidth]{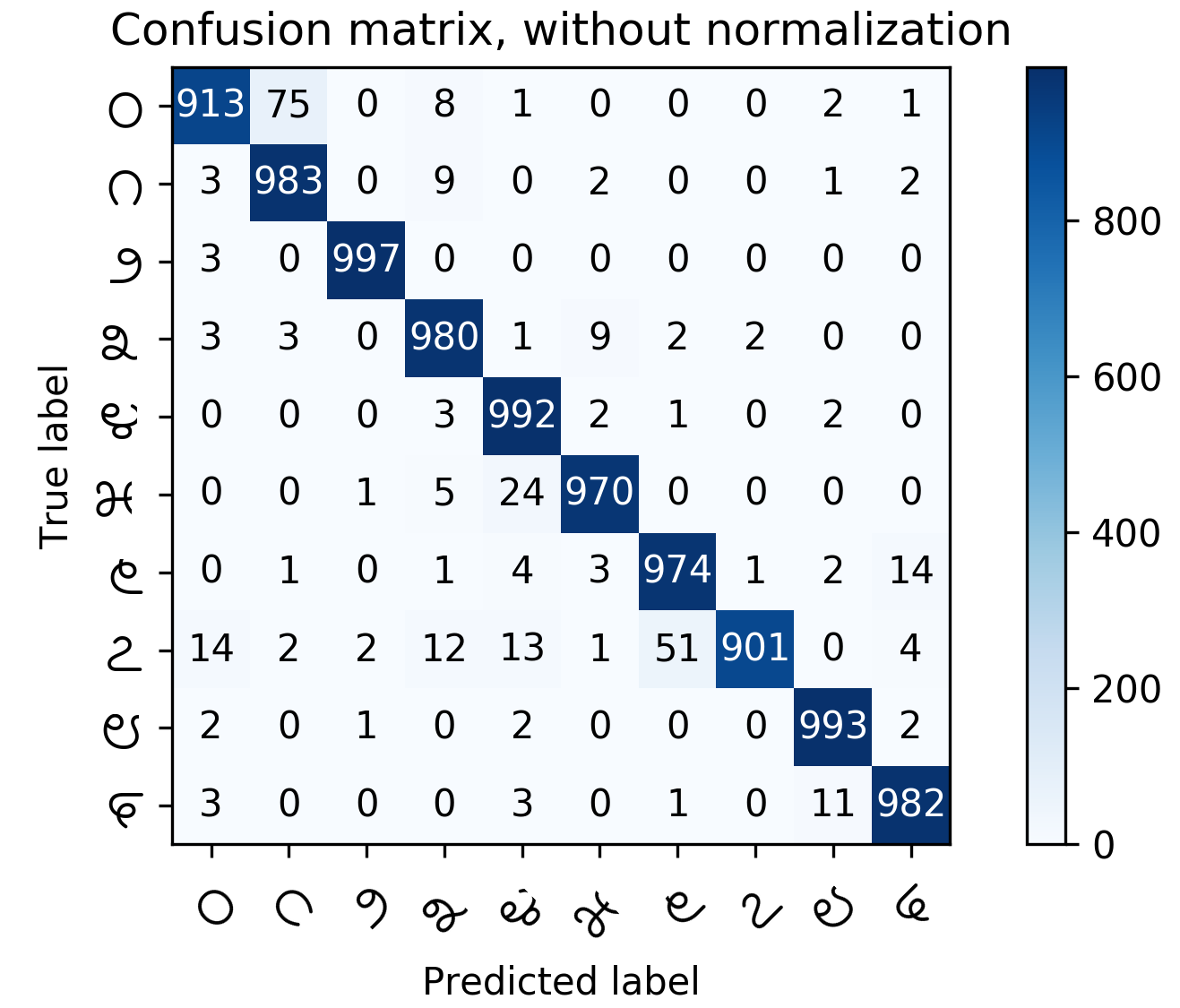}
\end{center}
   \caption{Confusion matrix with regards to the Kannada-MNIST datasets}
\label{fig:cmat_main}
\end{figure}
\begin{figure}[t]
\begin{center}
\includegraphics[width=0.8\linewidth]{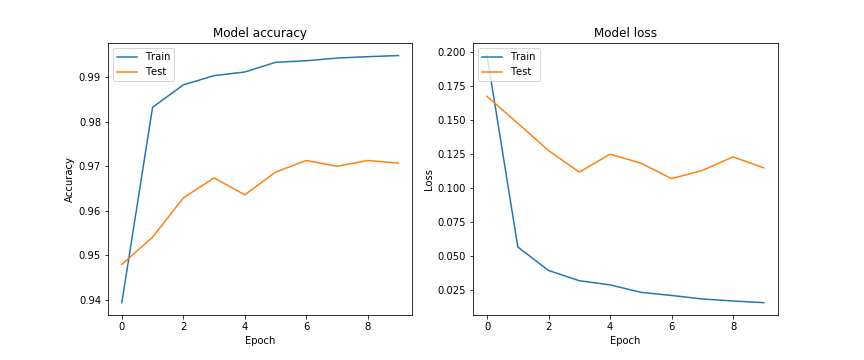}
\end{center}
   \caption{Train and test accuracies for the CNN trained and tested on the \textit{main} dataset }
\label{fig:main}
\end{figure}
\begin{figure}[t]
\begin{center}
\includegraphics[width=0.8\linewidth]{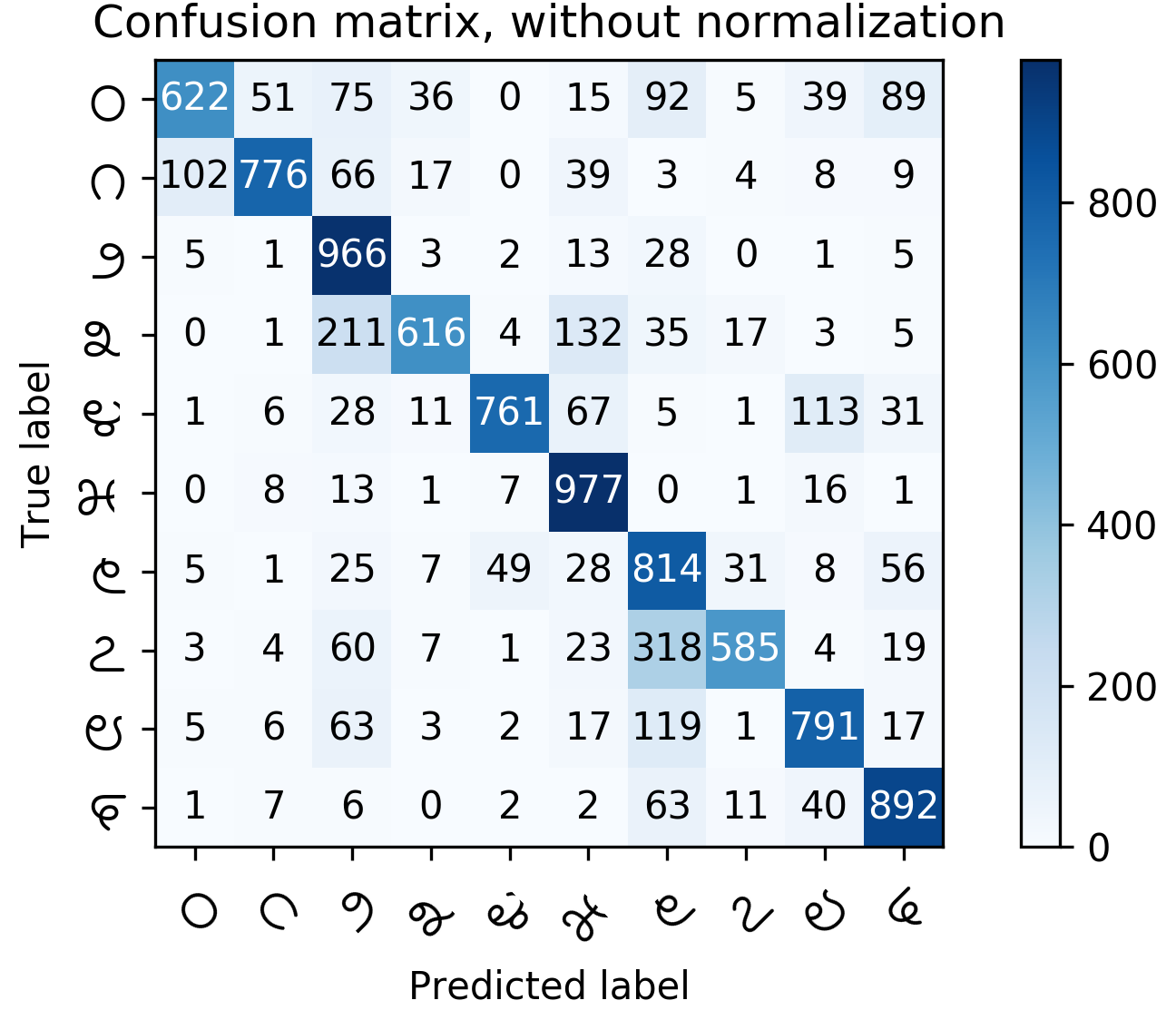}
\end{center}
   \caption{Confusion matrix for the \texttt{dig}-10k dataset}
\label{fig:cmat_dig}
\end{figure}

In this section, we present the classification results obtained by training an off-the-shelf CNN\footnote{\url{https://github.com/keras-team/keras/blob/master/examples/mnist_cnn.py}} (See Fig \ref{fig:cnn}) using Adadelta optimizer with \texttt{learning-rate=1.0} and$\rho=0.95$. For the \textit{main} dataset, with $60,000-10,000$ train-test split, we achieved $97.13\%$ top-1 accuracy. The classification report is as shown in Table \ref{tab:main} and the epoch-wise accuracy and loss plots are as shown in Fig \ref{fig:main}. In Fig \ref{fig:cmat_main}, we have the confusion matrix.
This pre-trained CNN achieved $76.2\%$ top-1 accuracy on the \texttt{dig}-10k dataset.  In terms of precision, class-2 ($63.9\%$) and class-6 ($55.1\%$) were the most challenging. In terms of recall, classes 0,3 and 7 were all at the sub-$61\%$ level. This showcases the fragile nature of the CNN's ability to truly generalize across author-cohorts and provides for an interesting challenge to the machine learning community at large. Fig \ref{fig:cmat_dig} and Table \ref{tab:dig} provide the confusion matrix and the per-class classifcation report for the \texttt{dig} dataset.

\begin{figure}[t]
\begin{center}
\includegraphics[width=0.8\linewidth]{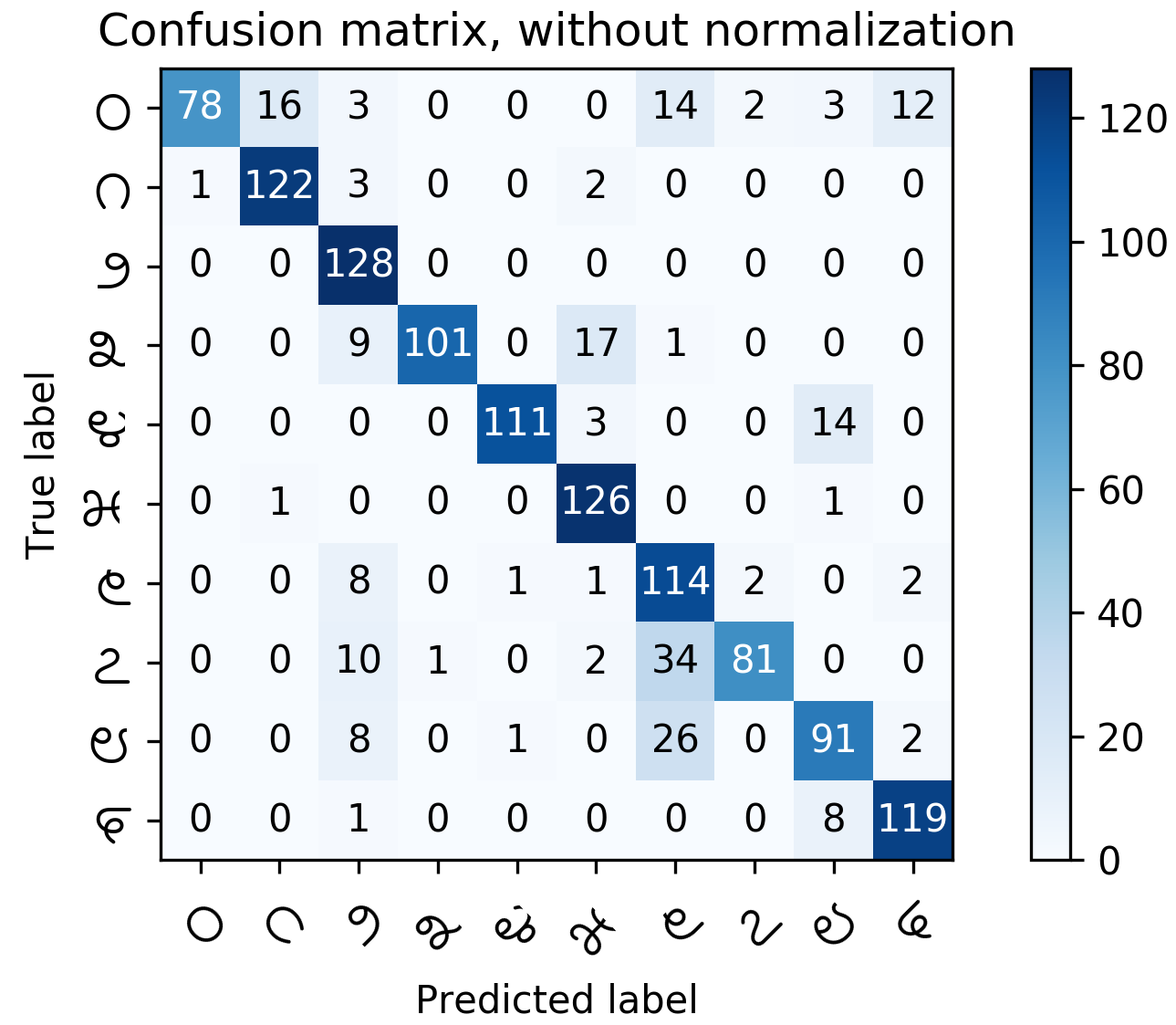}
\end{center}
   \caption{Un-normalized confusion matrix for the 1280 digits dataset using in \cite{prabhu2019fonts} }
\label{fig:cmat_iclr}
\end{figure}

%%%%%%%%%%%%%%%%%%%%%%%%%%%%%%
\begin{table}[]
\begin{center}
\begin{tabular}{lllll}
class         & precision & recall & f1-score & support \\
0             & 0.9702    & 0.9130 & 0.9408   & 1000    \\
1             & 0.9239    & 0.9830 & 0.9525   & 1000    \\
2             & 0.9960    & 0.9970 & 0.9965   & 1000    \\
3             & 0.9627    & 0.9800 & 0.9713   & 1000    \\
4             & 0.9538    & 0.9920 & 0.9725   & 1000    \\
5             & 0.9828    & 0.9700 & 0.9763   & 1000    \\
6             & 0.9466    & 0.9740 & 0.9601   & 1000    \\
7             & 0.9967    & 0.9010 & 0.9464   & 1000    \\
8             & 0.9822    & 0.9930 & 0.9876   & 1000    \\
9             & 0.9771    & 0.9820 & 0.9796   & 1000    \\
accuracy      &         &     & 0.9685   & 10000   \\
macro\_avg    & 0.9692    & 0.9685 & 0.9684   & 10000   \\
weighted\_avg & 0.9692    & 0.9685 & 0.9684   & 10000   \\
              &           &        &          &        
\end{tabular}
\end{center}
   \caption{Classification report for the Kannada MNIST dataset}
   \label{tab:main}
\end{table}

%%%%%%%%%%%%%%%%%%%%%%%%%%%%%%%%%

\begin{table}[]
\begin{center}
\begin{tabular}{lllll}
Class         & precision & recall & f1-score & support \\
0             & 0.8360    & 0.6074 & 0.7036   & 1024    \\
1             & 0.9013    & 0.7578 & 0.8233   & 1024    \\
2             & 0.6385    & 0.9434 & 0.7615   & 1024    \\
3             & 0.8787    & 0.6016 & 0.7142   & 1024    \\
4             & 0.9191    & 0.7432 & 0.8218   & 1024    \\
5             & 0.7441    & 0.9541 & 0.8361   & 1024    \\
6             & 0.5511    & 0.7949 & 0.6509   & 1024    \\
7             & 0.8918    & 0.5713 & 0.6964   & 1024    \\
8             & 0.7732    & 0.7725 & 0.7728   & 1024    \\
9             & 0.7936    & 0.8711 & 0.8305   & 1024    \\
accuracy      &          &       & 0.7617   & 10240   \\
macro\_avg    & 0.7927    & 0.7617 & 0.7611   & 10240   \\
weighted\_avg & 0.7927    & 0.7617 & 0.7611   & 10240   \\
              &           &        &          &        
\end{tabular}
\end{center}
   \caption{Classification report for the \texttt{dig} dataset}
   \label{tab:dig}
\end{table}

\begin{figure}[t]
\begin{center}
\includegraphics[width=0.8\linewidth]{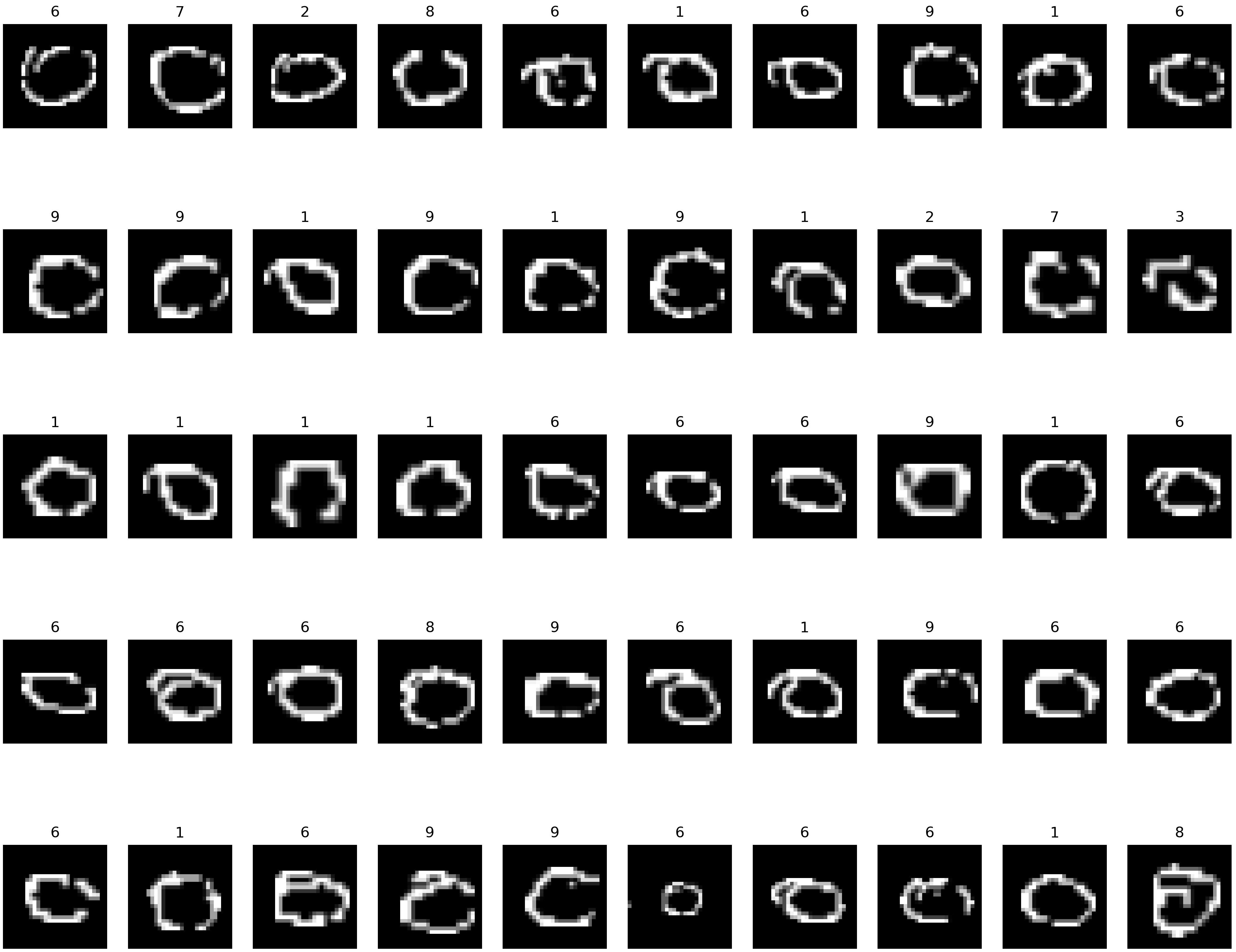}
\end{center}
  \caption{Images belonging to class-0 in the 1280-digits dataset that were misclassified by the CNN trained on the main dataset}
\label{fig:iclr_0}
\end{figure}

\begin{figure}[t]
\begin{center}
\includegraphics[width=0.8\linewidth]{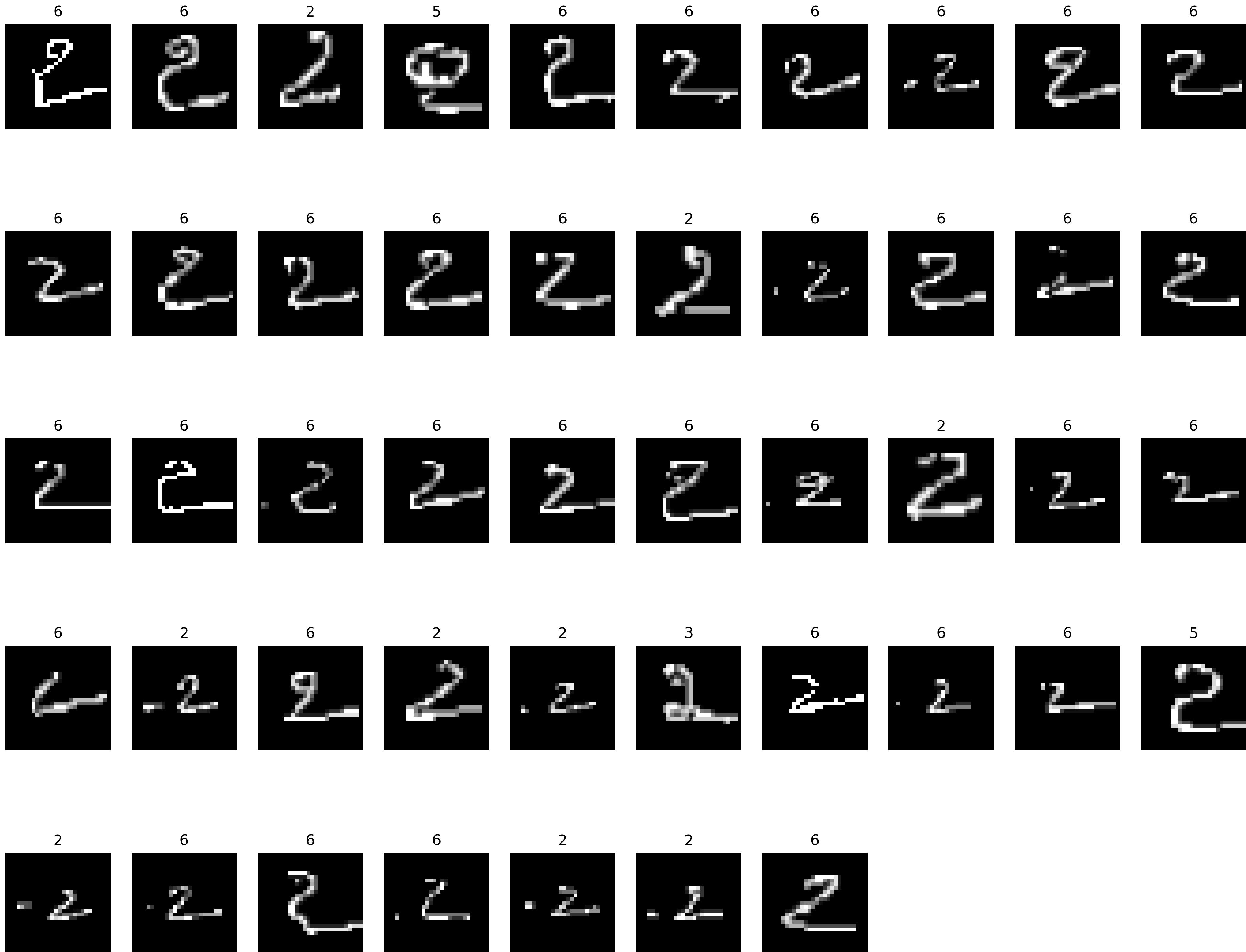}
\end{center}
  \caption{Images belonging to class-7 in the 1280-digits dataset that were misclassified by the CNN trained on the main dataset}
\label{fig:iclr_7}
\end{figure}

\begin{figure}[t]
\begin{center}
\includegraphics[width=0.8\linewidth]{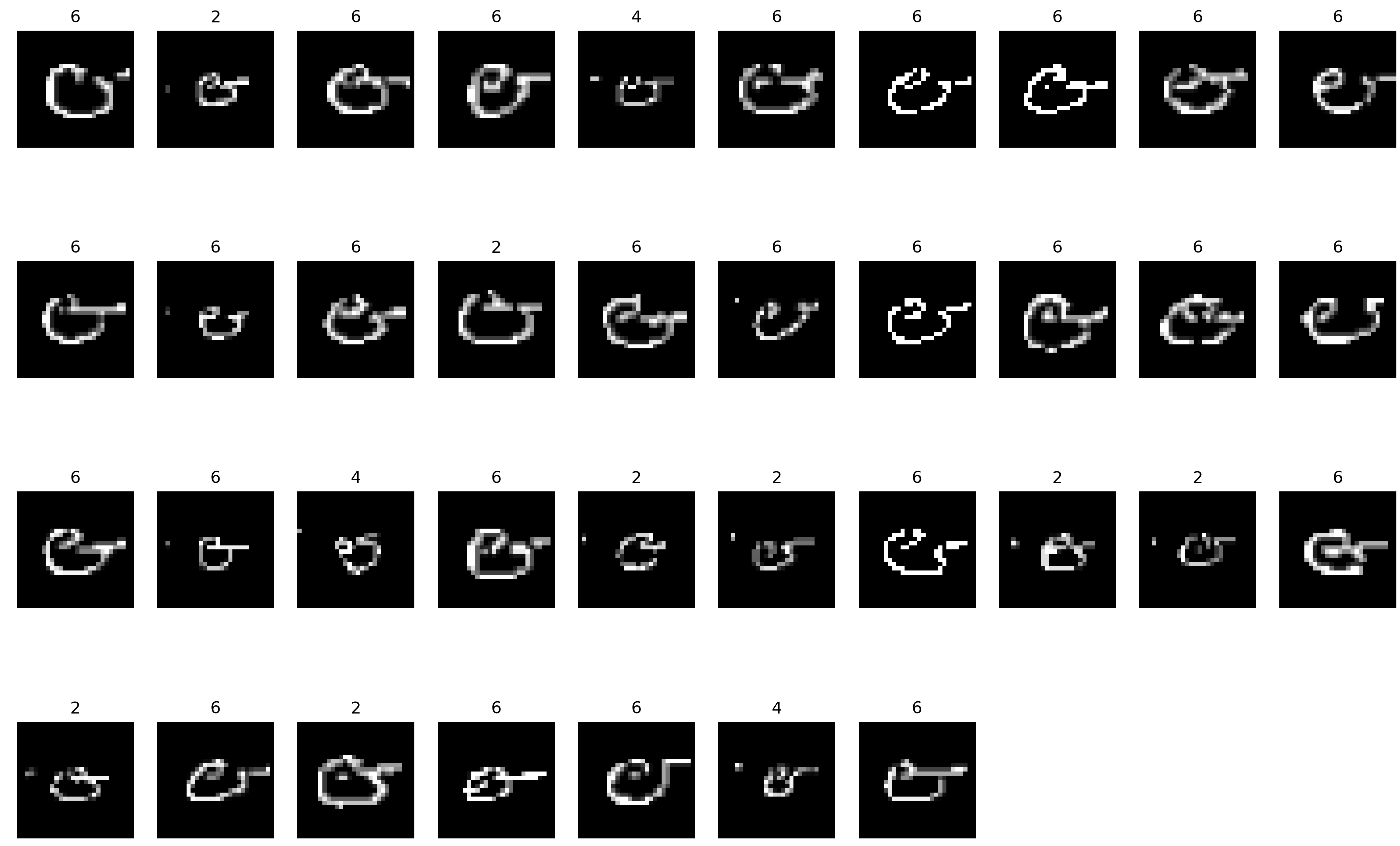}
\end{center}
  \caption{Images belonging to class-8 in the 1280-digits dataset that were misclassified by the CNN trained on the main dataset}
\label{fig:iclr_8}
\end{figure}

%%%%%%%%%%%%%%%%%%%%%%%%%%%%%%%%%%%%%%%%%
Lastly, for the single-author $1280$ digits dataset used in \cite{prabhu2019fonts}, we achieved $83.67\%$ top-1 accuracy. Again, as seen in Table \ref{tab:iclr} the CNN did struggle to achieve good precision class-6 ($60\%$) and good recall for classes -0 and 7 ($60-63\%$). Figure \ref{fig:cmat_iclr} provides the confusion matrix for the same. Given the smaller size of the test dataset, we did dig in to visualize the images of the digits that the CNN classified for classes 0 (Fig  \ref{fig:iclr_0}), 7(Fig  \ref{fig:iclr_7}) and 8(Fig  \ref{fig:iclr_8}). The title of each of the plots represents the predicted class by the CNN. As seen, for a human eye, most of the images do look like the glyphs. But, upon closer inspection, we observe the presence of \textit{rogue} non-glyph pixels (akin to naturally occurring adversarial perturbations) and discontinuities in the strokes in many of the erroneously classified images, which we posit might well explain the misclassifications.
\begin{table}[]
\begin{center}
\begin{tabular}{lllll}
Class         & Precision & Recall & f1-score & Support \\
0             & 0.99      & 0.61   & 0.75     & 128     \\
1             & 0.88      & 0.95   & 0.91     & 128     \\
2             & 0.75      & 1.00   & 0.86     & 128     \\
3             & 0.99      & 0.79   & 0.88     & 128     \\
4             & 0.98      & 0.87   & 0.92     & 128     \\
5             & 0.83      & 0.98   & 0.90     & 128     \\
6             & 0.60      & 0.89   & 0.72     & 128     \\
7             & 0.95      & 0.63   & 0.76     & 128     \\
8             & 0.78      & 0.71   & 0.74     & 128     \\
9             & 0.88      & 0.93   & 0.90     & 128     \\
Accuracy      &           &        & 0.84     & 1280    \\
macro\_avg    & 0.86      & 0.84   & 0.84     & 1280    \\
weighted\_avg & 0.86      & 0.84   & 0.84     & 1280    \\
\end{tabular}
\end{center}
   \caption{Classification report for the 1280 digits dataset used in \cite{prabhu2019fonts}}
   \label{tab:iclr}
\end{table}

\section{Conclusion and Future work}
\begin{figure}[ht]
\begin{center}
\includegraphics[width=0.8\linewidth]{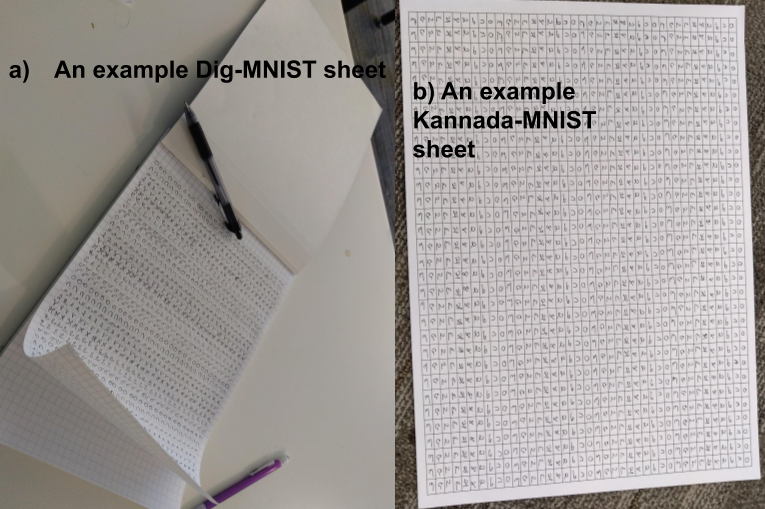}
\end{center}
   \caption{Photos of hard copies of the handwritten sheets for the two datasets}
\label{fig:hard_copies}
\end{figure}

In this paper, we described in detail the creation of a new handwritten digits dataset for the Kannada language, which we term as \texttt{Kannada-MNIST} dataset. We have duly open sourced all aspects of the dataset creation including the raw scan images, the specific brand of paper used\footnote{We have decided to also donate the hard copies of the handwritten digit-grids (See Fig \ref{fig:hard_copies}) to an academic research lab. If interested, the researcher(s) are requested to contact the main author with a proposal}, the exact scanner model used, the signal processing script used to slice and extract the individual digits and the CNN models used to obtain the baseline accuracies. We were able to attain $\sim 97\%$ top-1 accuracy when we trained and tested on what we term as the \textit{main} dataset with 60000 $28 \times 28$ gray-scale training images and 10000 test images. This is meant to be in a drop-in replacement for the standard MNIST dataset. We also achieved a top-1 accuracy of $\sim 77\%$ when we trained on the 60000 \textit{main} dataset and tested on $10240$ $28 \times 28$ gray-scale test images from what we term as the \texttt{Dig-MNIST} dataset. The images in the \textit{Dig}-MNSIT dataset are noisier with smudges and grid borders sneaking in during the grid-image segmentation phase. 
\\We propose the following open challenges to the machine learning community at large.
\begin{enumerate}
\item Achieve MNIST-level accuracy by training on the Kannada-MNIST dataset and testing on the Dig-MNIST dataset without resorting to image pre-processing.
\item To characterize the nature of catastrophic forgetting when a CNN pre-trained on MNIST is retrained with Kannada-MNIST. This is particularly interesting given the observation that the typographical glyphs for 3 and 7 in Kannada-MNIST hold uncanny resemblance with the glyph for 2 in MNIST.  
\item Get a model trained on purely synthetic data generated\footnote{\url{https://github.com/vinayprabhu/Kannada_MNIST/blob/master/colab_notebooks/5)_Synthetic_seed_image_generation.ipynb}} using the fonts (as in \cite{prabhu2019fonts}) and augmenting using frameworks such as \cite{castro2018morphomnist} and \cite{augmentor} to achieve high accuracy of the Kannada-MNIST and Dig-MNIST datasets.
\item Replicate the procedure described in the paper across different languages/scripts, especially the Indic scripts.
\item With regards to the \textit{dig}-MNIST dataset, we saw that some of the volunteers had transgressed the borders of the grid and hence some of the images either have only a partial slice of the glyph/stroke or have an appearance where it can be argued that they could potentially belong to either of two different classes. With regards to these images, it would be worthwhile to see if we can design a classifier that will allocate proportionate softmax masses to the \textit{candidate} classes.
\item The main reason behind us sharing the raw scan images was to foster research into auto-segmentation algorithms that will parse the individual digit images from the grid, which might in turn lead to higher quality of images in the upgraded versions of the dataset.
\end{enumerate}
\section*{Acknowledgement}
To begin with, we'd like to acknowledge the contribution of all the volunteers who contributed towards this dataset. Specifically, we'd like to thank Kaushik BK, who was instrumental in helping manage the cohort of 65 volunteers who contributed to the main dataset in Bangalore, India. We'd also like to acknowledge the contributors of the \texttt{Dig}-MNIST dataset in Redwood City, including John Whaley, Nick Richardson, Joseph Gardi and Preethi Sheshadri. Last but not the least, we'd like to acknowledge the helpful advice shared by the authors of the K-MNIST paper, Tarin Clanuwat, Alex Lamb(Mila) and David Ha (Google Brain).

\bibliographystyle{unsrt}  
\bibliography{references}
  %%% Remove comment to use the external .bib file (using bibtex).

\end{document}